\documentclass[letterpaper, 10 pt, conference]{ieeeconf}  %

\IEEEoverridecommandlockouts                              %

\usepackage{graphics} %
\usepackage{graphicx} %
\usepackage{subcaption,graphicx}

\usepackage{amsmath} %
\usepackage{booktabs}
\usepackage{hyperref}
\usepackage{fancyhdr}
\usepackage{multirow}
\usepackage{threeparttable}
\usepackage[table,dvipsnames]{xcolor}
\usepackage{breqn}
\usepackage{gensymb}
\usepackage{tabu}
\usepackage{arydshln}
\usepackage[super]{nth}
\usepackage{flushend}
\title{\LARGE \bf Condition-Invariant Multi-View Place Recognition}

\author{Jose M. Facil$^{1}$, Daniel Olid$^{2}$, Luis Montesano$^{1,3}$
and Javier Civera$^{1}$%
\thanks{$^{1}$ Jose M. Facil, Luis Montesano and Javier Civera are with the Robotics, Perception and Real Time Group, I3A, University of Zaragoza
        {\tt\small \{jmfacil,montesano,jcivera\}@unizar.es}}%
\thanks{$^{2}$ Daniel Olid is now at Opel Espa{\~n}a but worked in this paper earlier, while he was a student at the University of Zaragoza.
        }%
\thanks{$^{3}$ Luis Montesano is also with Bitbrain}%
\thanks{* This work was supported in part by the Spanish government (project DPI2015-67275) and in part by the Aragon regional government (Grupo DGA-T45\_17R/FSE), and by NVIDIA Corporation through the
donation of a Titan X and Xp GPUs.}
}

\begin{document}

\maketitle
\thispagestyle{empty}
\pagestyle{empty}

\begin{abstract}
    Visual place recognition is particularly challenging when places suffer changes that modify their appearance. Such changes are indeed common, e.g., due to weather, night/day, seasonal features or dynamic content. In this paper we leverage on recent place recognition research using deep networks; and explore how it can be improved by exploiting the information from multiple views. Specifically, we propose 3 different alternatives (Descriptor Grouping, Fusion and Recurrent Descriptors) for deep networks to combine visual features of several frames in a sequence. We show that our approaches produce more compact and better-performing descriptors than single- and multi-view baselines in the literature in two public databases. 
\end{abstract}
\section{Introduction}
Given a dataset of images taken at different places, visual place recognition \cite{lowry2016visual,arandjelovic2014dislocation,torii2013visual}  aims to identify the place of a new query image by associating it to one or several images of the dataset taken in the same location. Recent advances in computer vision have improved the performance of these algorithms, which are currently applied in several different applications such as image retrieval (\textit{e.g.}, \cite{noh2017large}), mapping and navigation in robotics \cite{pronobis2006discriminative,galvez2012bags,lee2019loosely}, autonomous driving \cite{mcmanus2014shady} and augmented reality (AR) \cite{middelberg2014scalable}. %

 One of the main challenges of visual place recognition is dealing with changes in the appearance of places \cite{garg2019semantic}. Indeed, place recognition is reasonably robust under small changes in viewpoint and illumination, due to the invariance of local features and rigidity checks \cite{galvez2012bags}. But, in constrast, non-rigid scene changes, wide baseline matching and extreme illumination variations are considerably more challenging and result in lower performance. 
 Using multiple frames in a sequence can improve the robustness of place recognition against such changes. But the sequence models proposed by the state of the art \cite{milford2012seqslam, naseer2018robust} are handcrafted for a certain set of assumptions (\textit{e.g.} overlapping trajectories, similar velocity patterns), and their performance suffers if they are not hold. Also, typically, they require a high number of frames. 
 
 Descriptors directly extracted from CNNs have shown good generalization properties \cite{gomez2015training}, but they usually do not exploit multi-view information. Improvements usually come at the cost of large descriptors (i.e. in the order of thousands or hundreds of thousands). The complexity of all place recognition algorithms depends on the size of the descriptor and the number of images in the database, the latest being typically high. This limits the applicability of these techniques in several applications as robotics and AR/VR, in which processing time is limited due to real-time constrained loops and limited on-board computational power.

\begin{figure}[t!]
    \centering
    \includegraphics[width=\linewidth]{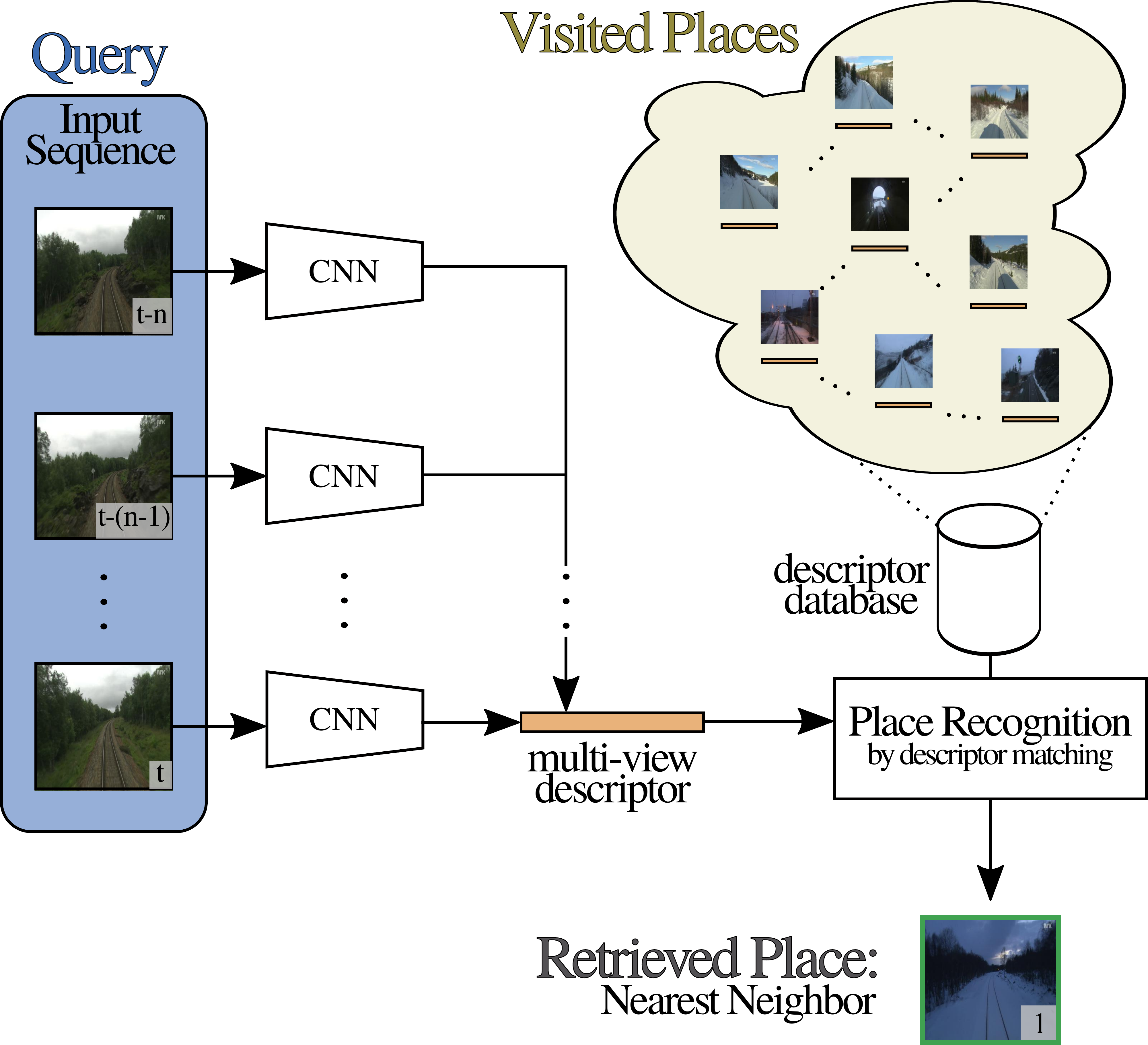}
    \caption{ \label{fig:teaser}Overview of our proposal. We extract descriptors (using deep networks) for small sequences of $n$ frames. We use such descriptors to find the closest match in a database of already visited places. }
   
\end{figure}

In this paper we target place recognition in the presence of challenging changes in the condition of an environment, that eventually happen in most of the scenes as time passes. For example day/night illumination, seasonal and weather changes, or objects that are moved (cars, persons or furniture). We propose and evaluate  three different deep network architectures that exploit multi-view and temporal information for place recognition: na{\"i}ve descriptor grouping, learning the fusion of single-view descriptors and recurrent networks using LSTM (Long Short Term Memory) layers \cite{hochreiter1997long}. An overview of our proposal can be seen in Fig. \ref{fig:teaser}. Up to our knowledge, ours are the first models that use deep learning to combine multi-view information for the purpose of place recognition.

We evaluated our models and compared it to state-of-the-art single-view deep models and a non-deep sequential one using two standard datasets: the Partitioned Nordland \cite{olid2018single} and Alderley \cite{milford2012seqslam}. 
The experimental results show that the performance of our three proposed multi-view models is better than single-view networks and that we also outperform SeqSLAM, a baseline for recognition from image sequences that does not use deep learning.  
Furthermore, our learned descriptors are at least one order of magnitude smaller than those of the state of the art showing that multi-view learning is able to extract relevant information for place recognition .

The rest of the paper is organized as follows. Section \ref{sec:rel} refers the related work. Section \ref{sec:models} gives the details of our network architectures, and section \ref{sec:training} of our     training. Finally, section \ref{sec:experiments} presents the experimental results and section \ref{sec:conclusions} the conclusions and lines for future work. Our \textbf{code} and a \textbf{video} showing our results can be found in our {project website}: \hbox{\small\url{http://webdiis.unizar.es/~jmfacil/cimvpr/}}. A reduced version of the video accompanies the paper as supplementary material. 

\newcommand{\modelone}{Descriptor Grouping}
\newcommand{\modeltwo}{Descriptor Fusion}
\newcommand{\modelthree}{Recurrent Descriptors}
\newcommand{\rulesep}{\unskip\ \vrule\ }
\begin{figure*}[ht!]
\begin{subfigure}[t]{0.32\textwidth}
    \centering
    \bf \large \modelone{}
\end{subfigure}
\begin{subfigure}[t]{0.32\textwidth}
    \centering
    \bf \large \modeltwo{}
\end{subfigure}
\begin{subfigure}[t]{0.32\textwidth}
    \centering
    \bf \large \modelthree{}
\end{subfigure}
\medskip
\vspace{0.1cm}
\medskip
\begin{subfigure}[t]{0.32\textwidth}
    \includegraphics[width=\textwidth]{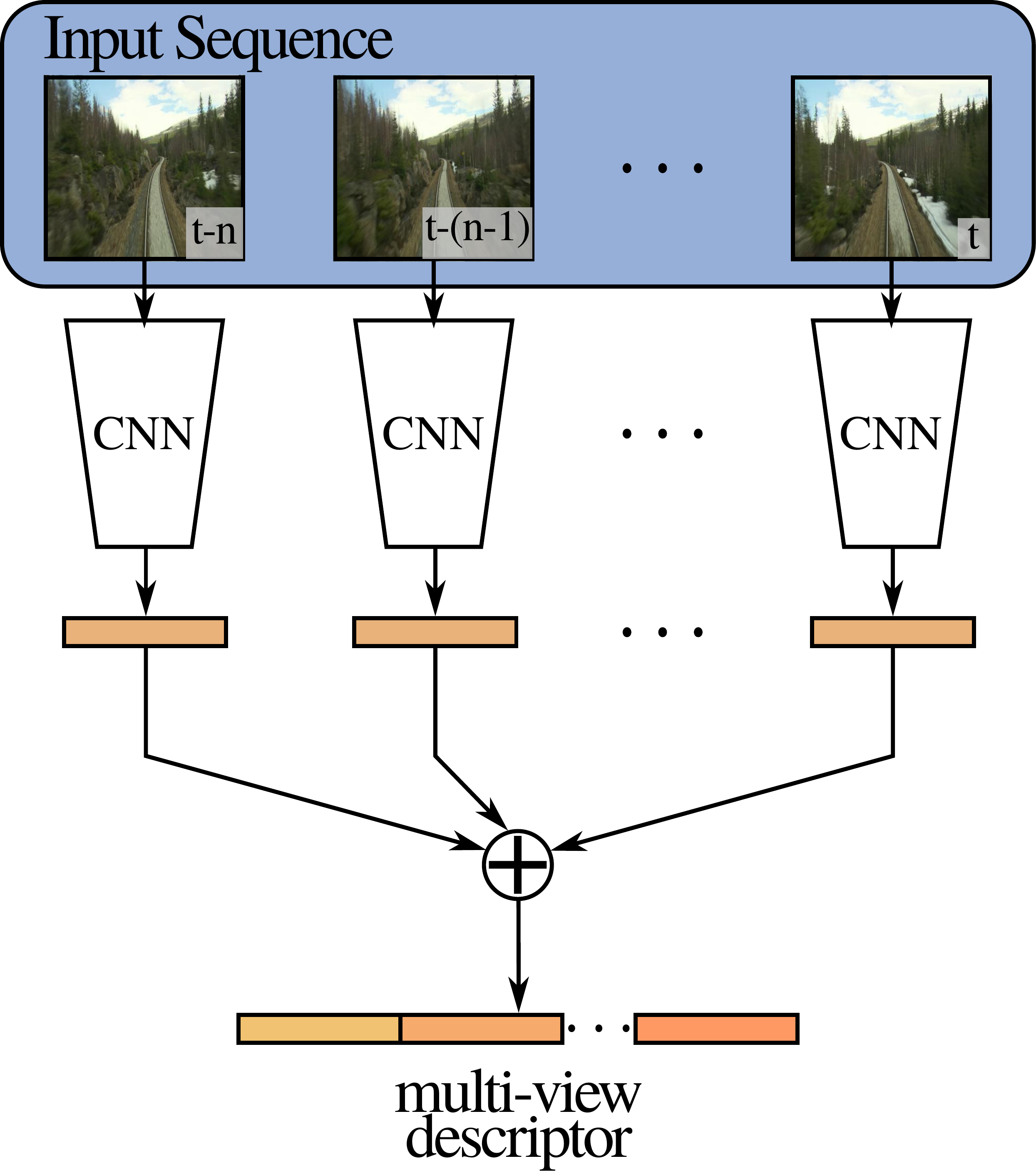}
   \subcaption{\label{fig:model1}}
\end{subfigure}
\rulesep
\begin{subfigure}[t]{0.32\textwidth}
    \includegraphics[width=\textwidth]{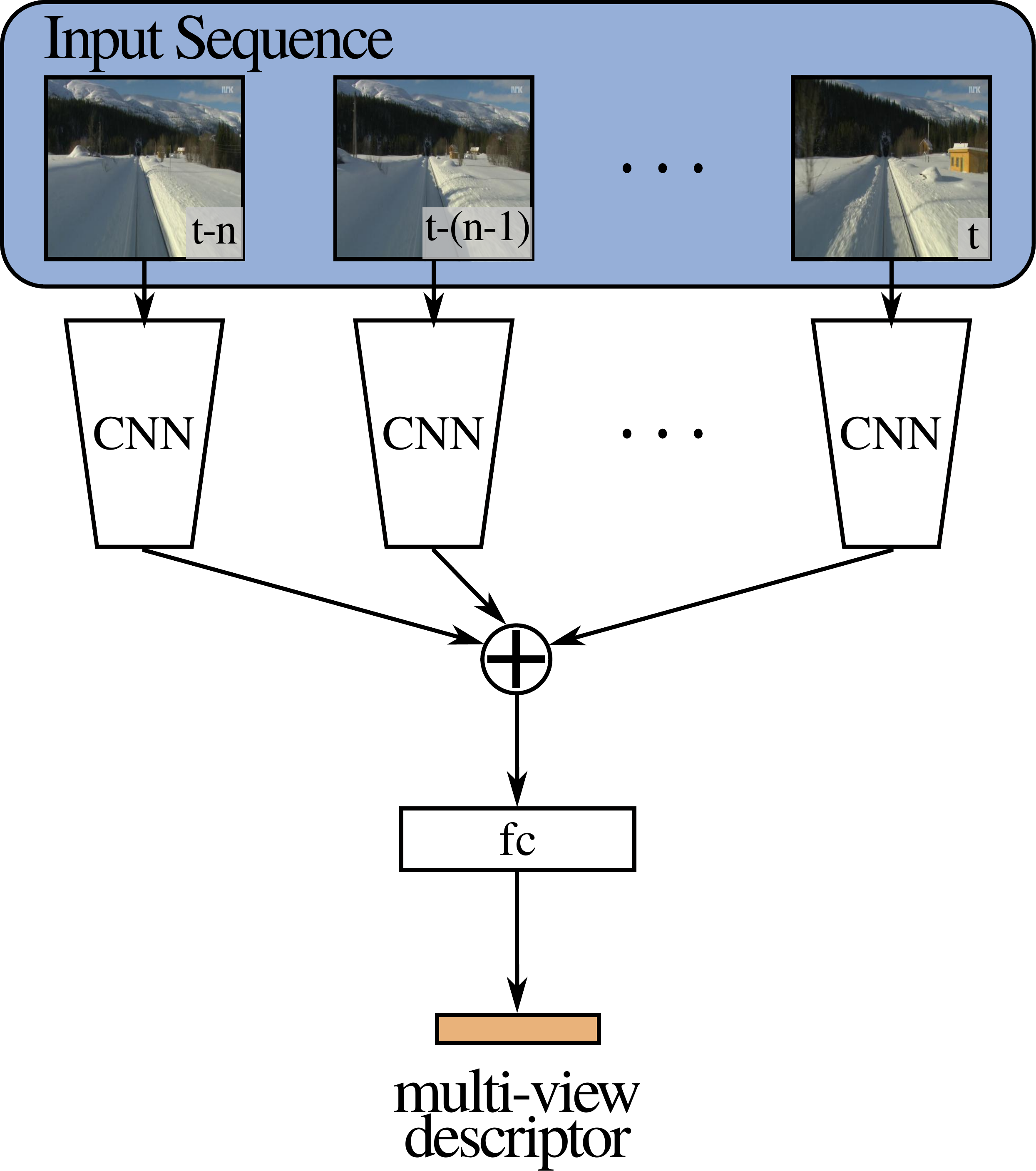}
    \subcaption{\label{fig:model2}}
\end{subfigure}
\rulesep
\begin{subfigure}[t]{0.32\textwidth}
    \includegraphics[width=\textwidth]{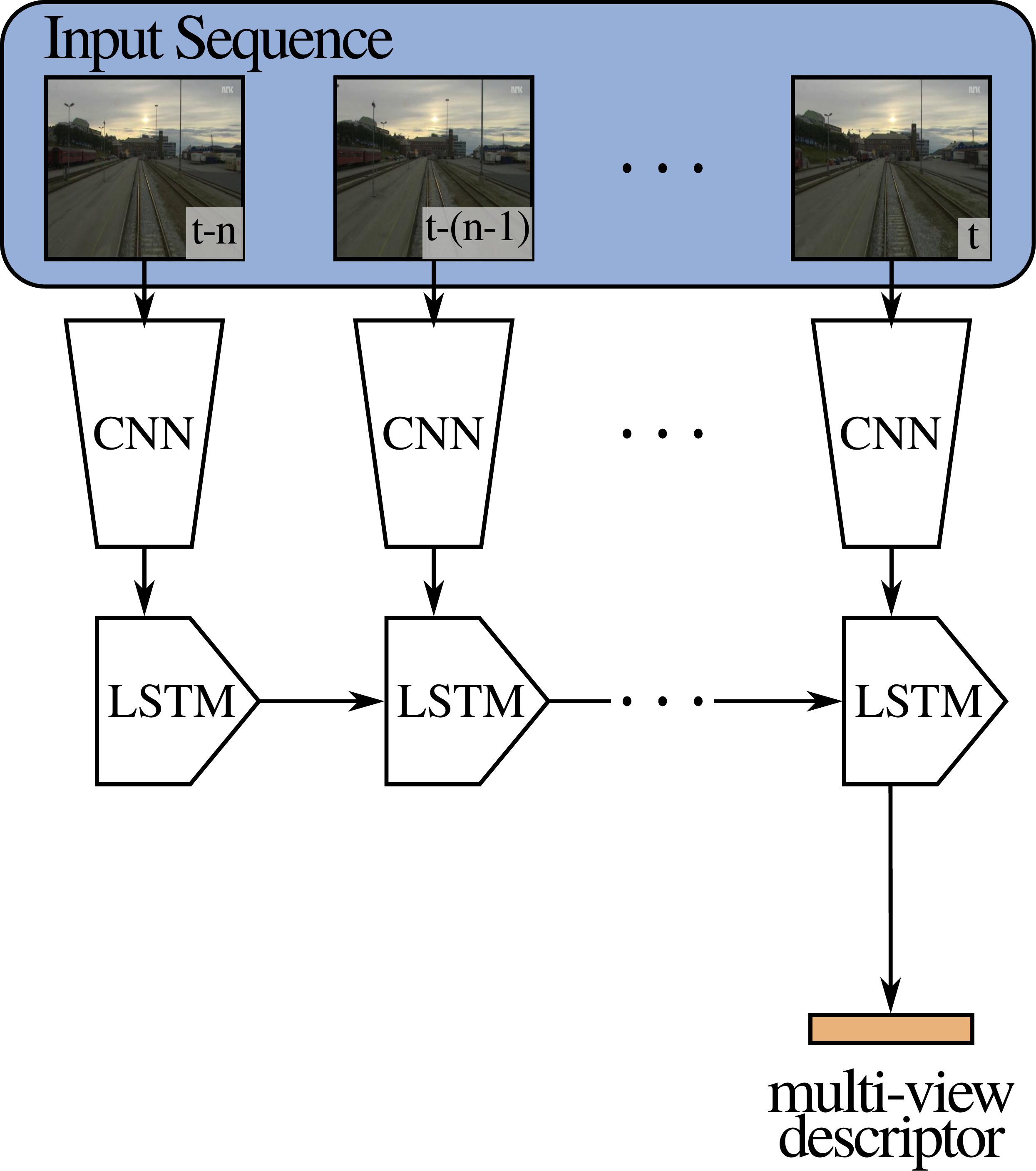}
    \subcaption{\label{fig:model3}}

\end{subfigure}
\caption{ \label{fig:model_cols}\textbf{Multi-view models} proposed in this paper. The sequence descriptors are used to retrieve a visited place via Nearest Neighbour. From left to right: (a) \textbf{\modelone{}}, where the descriptor of a sequence is the concatenation of all the single image descriptors. (b) \textbf{\modeltwo{}}, the output of the CNNs serves as inputo to a fully-connected layer that combines the information into a single descriptor. (c) \textbf{\modelthree{}}, the output of the CNNs serves as input to an LSTM network that integrates over time the single-image features to create a multi-image descriptor.}
\end{figure*}
\section{Related Work}
\label{sec:rel}

There have been many works addressing visual place recognition or related problems. For a general overview, we refer the reader to two surveys,  \cite{garcia2015vision} on topological mapping and  \cite{lowry2016visual} exclusively for visual place recognition. In this section we will focus on the most relevant to our work.

\subsection{Single-View Place Recognition}

Most of the work on visual place recognition extract descriptors from a single-frame. Some approaches have been based on handcrafted holistic image descriptors, like low-resolution thumbnails \cite{milford2012seqslam} or GIST descriptors \cite{murillo2013localization}. Although such approaches are very efficient, their performance degrades with large illumination and viewpoint changes and occlusions. Feature-based approaches (\textit{e.g.}, FAB-MAP \cite{cummins2008fab} and DBoW \cite{galvez2012bags}), relying on local information around salient points, are more robust to those changes. 

This type of descriptors are not robust to appearance changes due to scene dynamics, seasonal and weather changes, or extreme viewpoint or lighting variations. To address this, \cite{lowry2016supervised} used PCA to reduce the dimensionality of descriptors eliminating the dimensions that are influenced by condition changes. \cite{chen2018learning} incorporates attention in order to focus on the most relevant image features for place recognition. Lowry and Andreasson \cite{lowry2018lightweight} presented a model using SURF detector and HOG features and studies the use of Bag of Words and
Vectors of Locally Aggregated Descriptors (VLAD)
for place matching.

Descriptors based on CNNs have shown a high degree of robustness against appearance changes. S{\"u}nderhauf {\em et~al.} \cite{sunderhauf2015performance,sunderhauf2015place} showed that CNNs outperform other models, especially for drastic appearance changes. They used AlexNet \cite{krizhevsky2012imagenet}, pretrained on ImageNet \cite{russakovsky2015imagenet}. %
The features of AlexNet contain semantic information about the whole scene, which improves the invariance to certain appearance changes. Thereafter, may other works have studied CNNs as condition-invariant feature extractors \cite{gomez2015training,arandjelovic2016netvlad,arroyo2016fusion,chen2017deep,lopez2017appearance} and \cite{olid2018single}. G{\'o}mez-Ojeda {\em et~al.} \cite{gomez2015training} were the first that trained a network as single-image feature extractor for visual place recognition under appearance changes. In NetVLAD \cite{arandjelovic2016netvlad}, they proposed a new type of  layer inspired in VLAD, an
image representation commonly used in image retrieval.
Chen {\em et~al.} \cite{chen2017deep} proposed a network trained to classify the place the image was taken. Olid {\em et~al.} \cite{olid2018single} who built a model upon a pre-trained VGG-16 \cite{simonyan2014very} and fine-tuned it for the task in a Triplet-Siamese architecture. 

\subsection{Multi-View Place Recognition}

Although there are only a few works that attempt to consider temporal and multi-view information for place recognition, they all have shown that sequences provide extra information for place recognition. 
For instance, DBoW \cite{galvez2012bags} incorporates a  temporal consistency constraint. SeqSLAM \cite{milford2012seqslam} and following works (e.g., \cite{pepperell2014all}) use sequence matching, similarly to \cite{newman2006outdoor}. Differently to our approach, they  %
assume linear temporal correlation for sequence matching. We also use semantic information for the matching where they use downsampled images. 

More recently, \cite{naseer2018robust} used a graph of single-view descriptors (HOG and AlexNet-based) to model and match image sequences. Their approach is similar to SeqSLAM, with two main differences. The straightforward one is that they use different descriptors. The second one, more subtle, is that their search of the best-matching sequence does not assume a constant speed variation between the sequences. SeqSLAM looks for straight lines in the similarity matrix, while \cite{naseer2018robust} uses a more sophisticated model. In any case, none of them can addresses changes in the sequence direction. And also, they typically rely on long-term sequence matching (\textit{i.e.}, query and database sequences having many consecutive matching frames), which is not always the case.
Both sequence models are handcrafted and up to our knowledge there are no models that, as ours, learn to combine single-view CNN descriptors from data. %

\section{Network Architectures}
\label{sec:models}

In this section, we discuss four different models: A single-view one for place recognition, based on ResNet-50, and three different extensions for multi-view place recognition. 

\subsection{Single-View ResNet-50}
Our first network is based on the model presented in \cite{olid2018single}. The main difference is that we start from ResNet-50 \cite{he2016deep} pretrained on ImageNet \cite{russakovsky2015imagenet} as our backbone, instead of VGG-16 \cite{simonyan2014very}. Although it is common to directly use the descriptors of different layers (see \ref{sec:expnordland} for results on this), in our case we added and trained a fully connected layer after ResNet-50 to learn a $128$-dimensional descriptor especially designated to the task of visual place recognition. We chose a size of $128$ experimentally, as a good compromise between performance and compacity.

\subsection{\modelone{}}
In order to include temporal information into the descriptors, our first approach is the  na{\"i}ve concatenation of the descriptors of consecutive frames, see Fig. \ref{fig:model1}.  Thus, starting from our previous single-view model, first we choose a window of frames ($n$) to work with, second we generate individual $128$ descriptors and third concatenate them. Concluding, as shown in the figure, with a $128\times n$ descriptor for the sequence. Notice that this model is trained only from single-view samples. Hence, the relation between consecutive frames is not learn and this model only provides a filtering effect.

\subsection{\modeltwo{}}
\modelone{}, as the simplest strategy to consider several frames, is limited in its capability to weight differently certain features (\textit{i.e.} features of some of the frames may be more representative of the place than others). For that reason we wanted to build a model that learned to fuse the information of our $n$-frames window into a more discriminant --as well as smaller-- $128$-dimensional descriptor. With this \modeltwo{} strategy, we add an extra fully connected layer that learns how to combine the outputs of $n$ ResNet-50 into a single compact descriptor. See Fig. \ref{fig:model2} for an illustration of this approach. As this network is able to learn how to weight the features from different frames, it can model more complex cases. For example, when sequences are recorded in reverse order \modelone{} is limited, while \modeltwo{} has the capability of learning a suitable fusion.

\subsection{\modelthree{}}
\modeltwo{} does not explicitly exploits the sequential nature of the data. In this model we wanted to update online the sequence descriptor as new frames come, keeping the most relevant previous information. With that intention, we propose a Recurrent Neural Network (see \modelthree{} in Fig. \ref{fig:model3}). In this model, every frame is the input to a ResNet-50, and the top layers serve as the input of a LSTM network \cite{hochreiter1997long}, that generates a $128$-dimensional descriptor. LSTMs keep an inner state, that is updated with each input frame, and the output depends on the state and the input. Differently to previous models, keeping a recurrent inner state allows this network to produce a descriptor from the first frame, and update it sequentially as more frames arrive.

\definecolor{queryorange}{RGB}{218, 124, 48}
\definecolor{placeblue}{RGB}{57, 106, 177}
\newcommand{\quseq}[1]{{\color{queryorange}\textit{query-sequence #1}}}

\newcommand{\place}[1]{{\color{placeblue}\textit{place #1}}}
\newcommand{\quseqe}{{\color{queryorange}\textit{query-sequence}}}

\newcommand{\placee}{{\color{placeblue}\textit{place}}}

\section{Training}
\label{sec:training}

\subsection{Convention for Same Place}

\begin{figure}[t!]
    \centering
    \includegraphics[width=\linewidth]{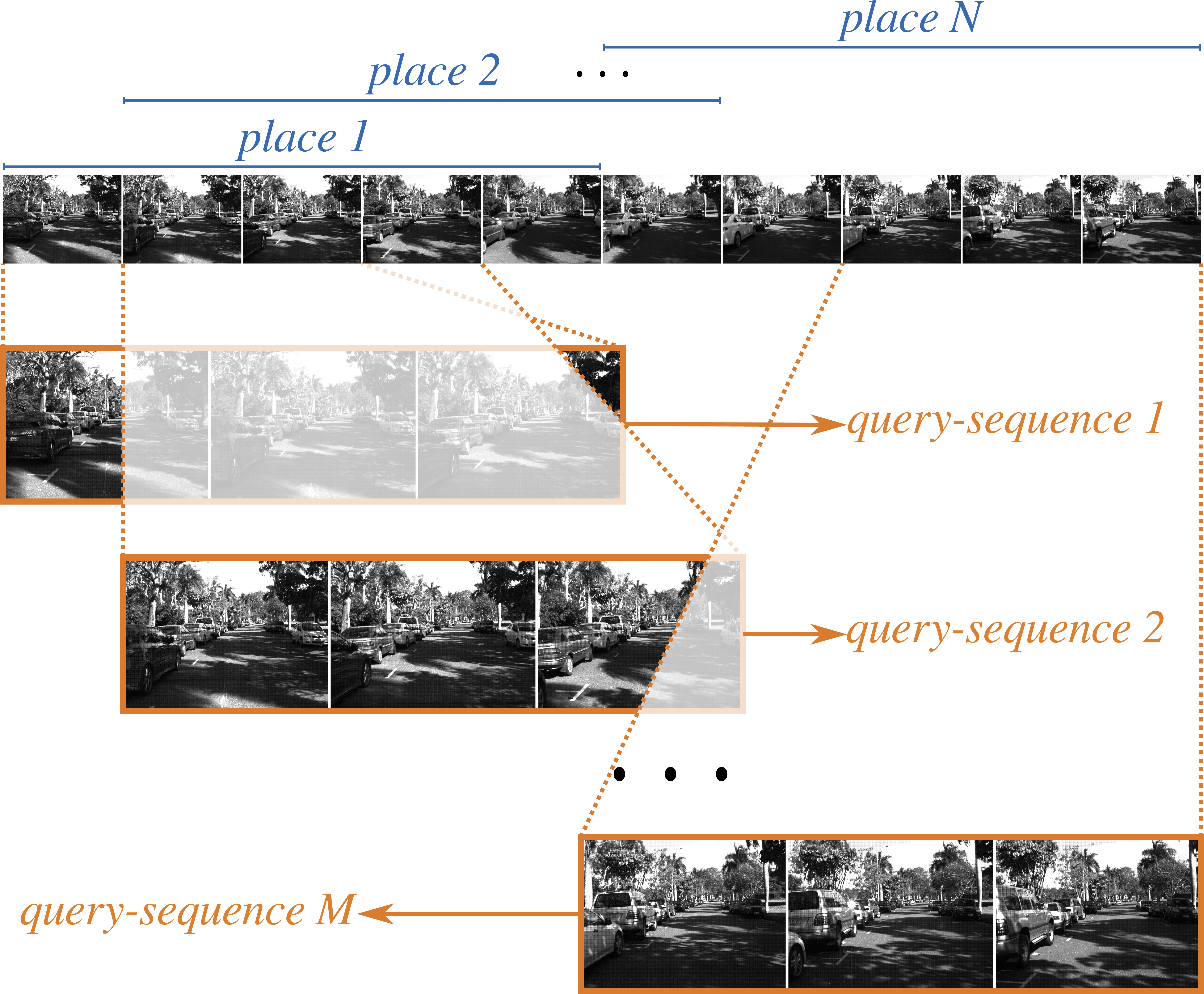}
    \caption{ \label{fig:places}Same place convention, illustrated with an example where the \quseqe{} has a length of 3 frames. A \placee{} represents a set of frames that are considered to be on the same place. Notice that a frame can be in more that one \placee{}. A \quseqe{} is an input sequence for our model. We want to recover the corresponding place of a \quseqe{}.}
\end{figure}

Since our descriptor is generated from a sequence of images (\quseqe{}) instead of a single image we must define when two \quseqe{} of $n$ frames are considered to be at the same \placee{} (the definition of a \placee{} being dataset-dependent). To illustrate this definition we will make use of Fig. \ref{fig:places}. The figure shows a sequence of frames and several examples of \quseqe{}, and also shows the set of frames that we consider as the same \placee{}. Therefore, during training, we consider two \quseqe{} to be on the same \placee{} if they contain two frames (one per \quseqe{}) that belong to the same \placee{}{}. For instance, in Fig. \ref{fig:places}, \quseq{1} and \quseq{2} belong to the same place, as the first frame of \quseq{1} belongs to \place{1}, the same as the first frame of \quseq{2}).

\subsection{Model training}

\begin{figure}[t!]
    \centering
    \includegraphics[width=\linewidth]{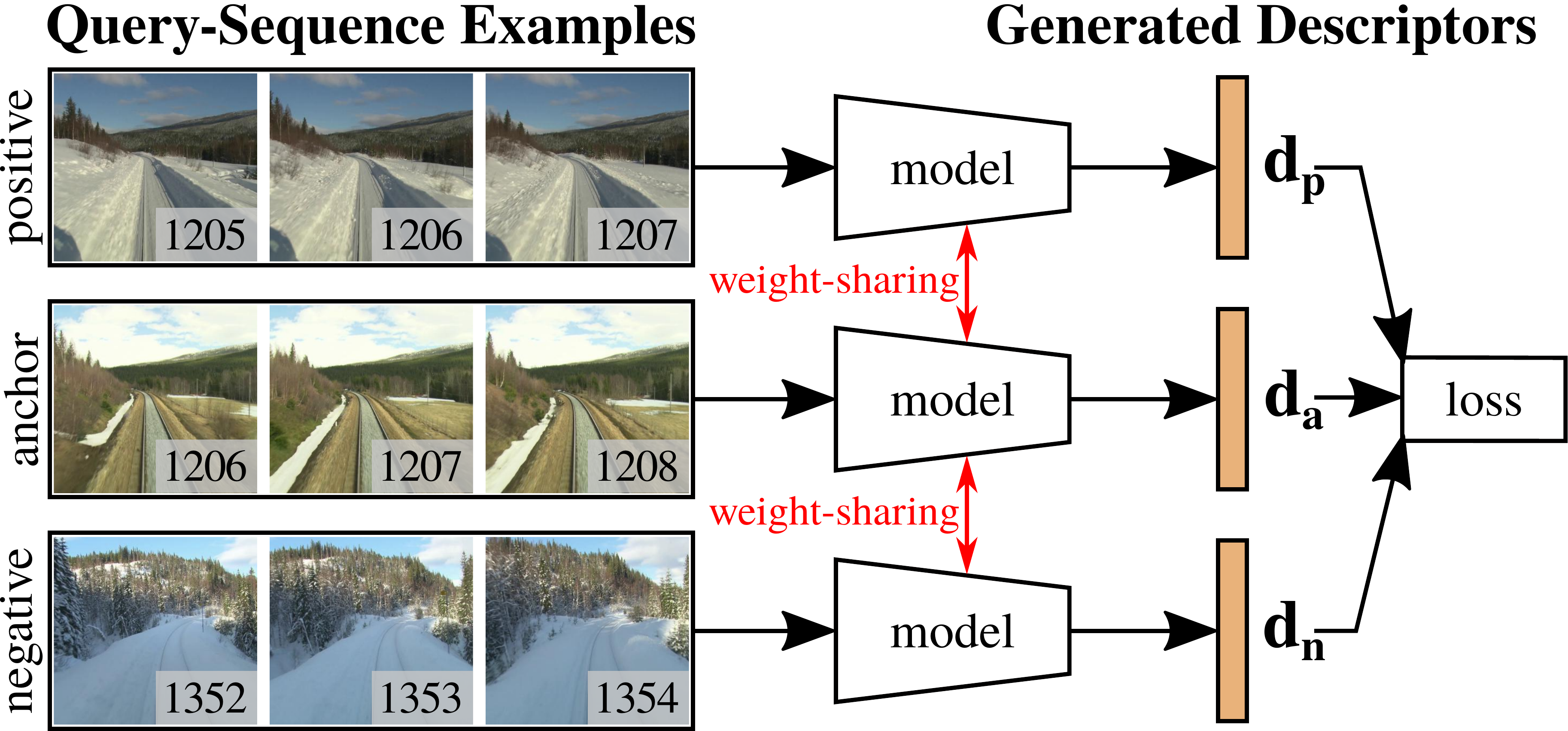}
    \caption{ \label{fig:triplet}Triplet architecture. We used this scheme for training all our models.}
\end{figure}

We start from ResNet-50 pre-trained on ImageNet in a standard classification task. We add the extra layers, and train them on our place-recognition task in our datasets. We trained all the models proposed in this work using a triplet architecture (see Fig. \ref{fig:triplet} for a scheme and \cite{olid2018single,gomez2015training} for more details). In a few words, triplet architectures are given 3 training samples: An anchor, a positive example and a negative one. During training, the objective is to reduce the distance between the anchor and positive descriptors, and to increase the distance between the anchor and the negative one. The loss we use to achieve that is the \textit{Wohlhart-Lepetit} loss \cite{wohlhart2015learning},
\begin{equation}
    \mathcal{L}=\max\left\{0, 1-\frac{||\mathbf{d_a}-\mathbf{d_n}||}{m+||\mathbf{d_a}-\mathbf{d_p}||}\right\},
\end{equation}
where $m$ (margin) is a parameter that limits the difference between the distances,  $\mathbf{d_a}$ is the descriptor generated for the \textit{anchor} image, $\mathbf{d_p}$ is the descriptor for the \textit{positive} sample and $\mathbf{d_a}$ is the descriptor for the \textit{negative} sample (see Fig. \ref{fig:triplet}). The specific training details for each model are as follows.\\
\subsubsection{\modelone{}} This model is trained as a single-view place recognition model. Hence, the data triplets consist of single images (an anchor image, a positive and a negative examples). During training, every single image generates a 128-dimensional descriptor, which means a \hbox{$128 \times n$} elements descriptor during test.
\subsubsection{\modeltwo{}} Our second model learns a fusion of features for an image sequence (\quseqe{}). Therefor, we concatenate the output of the ResNet-50, for each image, and add a fully-connected extra layer to generate a single descriptor of 128 elements. We train this model generating triples samples of $n$-frames \quseqe{}.
\subsubsection{\modelthree{}} In our last model we make a sequential update of the image descriptors using Recurrent Neural Networks, concretely an LSTM layer. In order to force the network to learn from the three images instead of only the last one, we add some random sampling in one of the $n$ images of the \quseqe{} plus a Dropout on the LSTM layer.

\section{Experimental Results}
\label{sec:experiments}
In this section we evaluate the three proposed models  on two datasets: the Partitioned Norland \cite{olid2018single} and Alderley \cite{milford2012seqslam}; and we compare them  with state-of-the-art single-view and sequence-based methods.

\noindent\textbf{Experimental setup:} For every method, we retrieve the single nearest neighbor as the matched place for a query image or sequence. We treat it as a correct match if it fits with the same-place convention for the dataset  
(\textit{i.e.} each dataset has its own ground-truth frame correspondence).
During the experiments, we set the \textbf{\quseqe{} length to 3 frames} for all our multi-view models. We observed in our experiments that, for more than 3 frames, the performance did not improved much. %

To compare different models we compute the precision of the model when recall is equal to 1. Hence, we retrieve a place for every query (the nearest neighbor) and compute the fraction of correct matches over the total number of queries.

 \subsection{Partitioned Nordland Dataset}
\label{sec:expnordland}
 Our experiments use the train-test split proposed by Olid {\em et~al.} \cite{olid2018single}. We trained our model with $24.5K$ images and evaluated its performance in a $3450$-images set. For every model we fixed the ResNet-50 parameters and trained the additional layers. We evaluated the performance of the descriptors of different layers of ResNet-50 and the best performing features are those of the layer  \textit{bn3d-branch2b} (3d-2b in the table), so we use these in our experiments. %
We train for $5$ full epochs, where each epoch corresponds to $840K$ triplet examples.

\begin{table}
\centering

\begin{tabular}{ccccc}
\textbf{Method} & \multicolumn{1}{p{1cm}}{\centering \textbf{Number\\of\\frames}}&\multicolumn{1}{p{1.1cm}}{\centering\textbf{Descriptor\\Size}}& \multicolumn{1}{p{1.1cm}}{\centering\textbf{Accuracy\\W vs S}} & \multicolumn{1}{p{1.1cm}}{\centering\textbf{Accuracy\\S vs W}} \\
\hline
    & $\#$& & $\%$ & $\%$ \\
    VGG16(pool4)& $1$& 100352& 51\% &  21\%\\
    VGG16(pool5)& $1$& 25088& 13\% &  7\%\\
    VGG16(fc6)& $1$& 4096& 6\% &  3\%\\
    VGG16(fc7)& $1$& 4096& 4\% &  3\%\\
    ResNet-50(3a-2a)& $1$& 100352& 42\% &  32\%\\
    ResNet-50(3d-2b)& $1$& 100352& 73\% &  42\%\\
    ResNet-50(4a-2a)& $1$& 50176& 62\% &  41\%\\
    ResNet-50(4b-2a)& $1$& 50176& 62\% &  31\%\\
    ResNet-50(4c-2a)& $1$& 50176& 50\% &  40\%\\
    ResNet-50(4f-2b)& $1$& 50176& 12\% &  8\%\\
    ResNet-50(5a-2a)& $1$& 100352& 43\% &  24\%\\
    Hybridnet \cite{chen2017deep} & $1$& 4096& 77\% &  $41\%$\\
    Amosnet \cite{chen2017deep}& $1$&4096 & 69\% &  48\%\\
    Lowry {\em et~al.} \cite{lowry2016supervised} & $1$&1860 & 67\% &  66\%\\
    Olid {\em et~al.} \cite{olid2018single} & $1$& 128& 75\% &  79\%\\
    \textbf{ours (single-view)} & $1$& 128 & 77\% &  75\%\\
    \hline
    Seqslam \cite{milford2012seqslam} & $3$&6144 & 31\% &  33\%\\
    Seqslam \cite{milford2012seqslam} & $10$& 20480& 71\% &  70\%\\
    Seqslam \cite{milford2012seqslam} & $100$& 204800 & \textbf{95\%} &  \textbf{94\%}\\
    \textbf{ours (grouping)} & $3$& 384 & \textbf{92\%} &  \textbf{92\%}\\
    \textbf{ours (fusion)} & $3$& 128 & 87\% &  86\%\\
    \textbf{ours (recurrent)} & $3$& 128 & 85\% &  86\%\\
    \cline{4-5}
  \multicolumn{3}{c}{}&\multicolumn{2}{c}{the bigger the better}
  
\end{tabular}
\caption{\label{tab:norland} Results on the \textbf{Partitioned Nordland Dataset} \cite{olid2018single}.
\textbf{\nth{1} column:} Method. \textbf{\nth{2} column:} Number of frames used for recognition (\textit{e.g.}, 1 stands for single-view). \textbf{\nth{3} column:} Descriptor size in 32-bits floating point numbers. \textbf{ \nth{4} column:} Winter vs Summer, query taken from winter and matched to summer database. \textbf{\nth{5} column:} Summer vs Winter, query taken from summer and matched to winter database. }
\end{table}

\noindent\textbf{Quantitative Results:} Results on Table \ref{tab:norland} show the performance of different models for visual place recognition on the Partitioned Nordland. In this table we report the hardest recognition cases, which are representative for the rest, specifically using the seasons Winter and Summer both as query and database respectively. The upper part of the table shows single-view models and the lower part shows the multi-view models. 

Our three proposals \textbf{ours (grouping)}, \textbf{ours (fusion)} and \textbf{ours (recurrent)} outperform very clearly our single-view approach \textbf{ours (single-view)}. Notice that they also outperform the state-of-the-art baselines. Among our multi-view proposals, \textbf{ours (grouping)} is the one achieving the best performance ($92\%$ of precision using $3$ frames). Notice that its descriptor size, $384$, is smaller than most of the single-view and mult-view baselines.

Out methods also outperform SeqSLAM \cite{milford2012seqslam}, a state-of-the-art baseline able to model information from several frames, when both use the same number of frames (specifically, $3$). As all multi-view approaches improve their performance when increasing the number of frames, we increased the number of frames used by SeqSLAM. Notice that, in order to outperform our approach, the number of frames has to be increased up to $100$ (with a descriptor size of 204800).

\begin{figure}
    \begin{subfigure}[t]{0.06\linewidth}
        \centering
    \end{subfigure}
    \begin{subfigure}[t]{0.32\linewidth}
        \centering
        \bf \modelone{}
    \end{subfigure}
    \begin{subfigure}[t]{0.30\linewidth}
        \centering
        \bf \modeltwo{}
    \end{subfigure}
    \begin{subfigure}[t]{0.32\linewidth}
        \centering
        \bf \modelthree{}
    \end{subfigure}
    \medskip
    \begin{subfigure}[t]{1\linewidth}
    \centering
    \includegraphics[width=\linewidth]{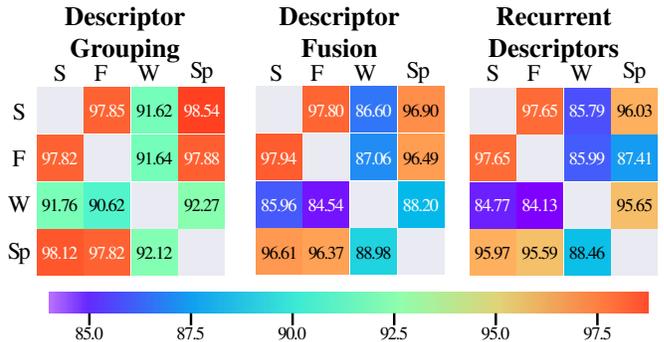}
    \end{subfigure}
    \caption{\label{fig:confusion}\textbf{Precision} on Partitioned Nordland Dataset \cite{olid2018single}. We evaluate the fraction of correct matches between all the seasons. \textbf{S} stands for summer, \textbf{F} for Fall, \textbf{W} for Winter and \textbf{Sp} for Spring. Notice that our best performing model, \modelone{}, never drops under $90\%$ of correct matches.}
    
\end{figure}

Fig. \ref{fig:confusion} shows the results of all our proposals for all query-reference combinations. As mentioned before, winter is always the hardest case. Notice, however, that none of our models ever drops under $80\%$ performance.

\newcommand{\experimentone}{Reverse Gear}
\newcommand{\experimenttwo}{Random Speed}
\definecolor{greenbox}{HTML}{00CE84} %
\newcommand{\rulesepexp}{\unskip\ \hrule\ }

\begin{figure}[t!]

\includegraphics[width=\linewidth]{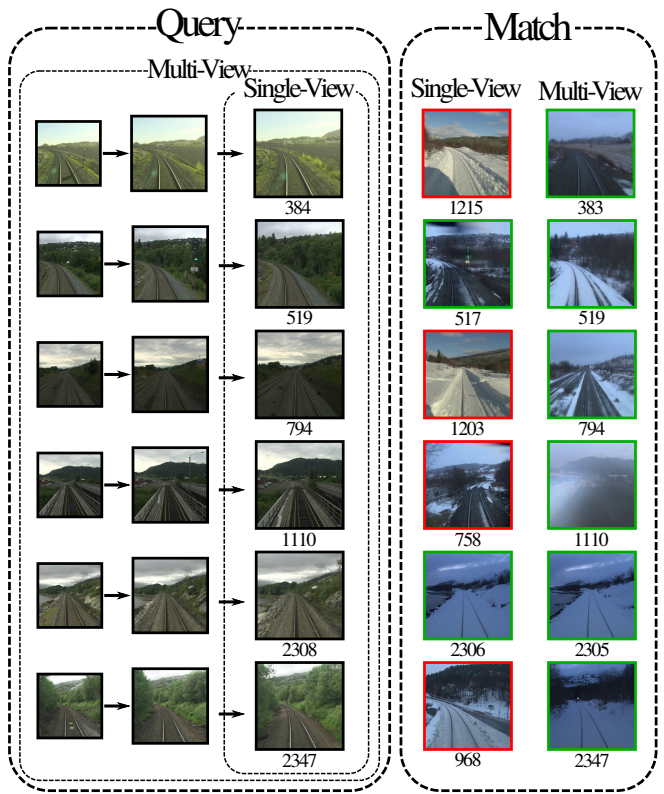}

\caption{ \label{fig:pnord_qual}\textbf{Example of Matched Places} for Single and Multi-View grouping model in the Partitioned Nordland Dataset \cite{olid2018single}. The retrieved image is framed on {\color{green} green} if it is a correct match or {\color{red}red} if it is incorrect. Mismatched frames are very similar and could even fool humans if they are not carefully inspected.
}

\end{figure}
\noindent\textbf{Qualitative Results:} Fig. \ref{fig:pnord_qual} shows some examples of matched places with the grouping model and illustrates when multi-view methods achieve better performance. Notice that, although our single-view method fails in these examples, some of the places are indeed very similar and would be hard to match even by humans.

\definecolor{cNT}{HTML}{EB4C42}%
\definecolor{cRG}{HTML}{1E90FF} %
\definecolor{cRS}{HTML}{FF8C00}%
\definecolor{cTOT}{HTML}{03C03C} %
\definecolor{greenbox}{HTML}{00CE84} %
\begin{figure}[t!]
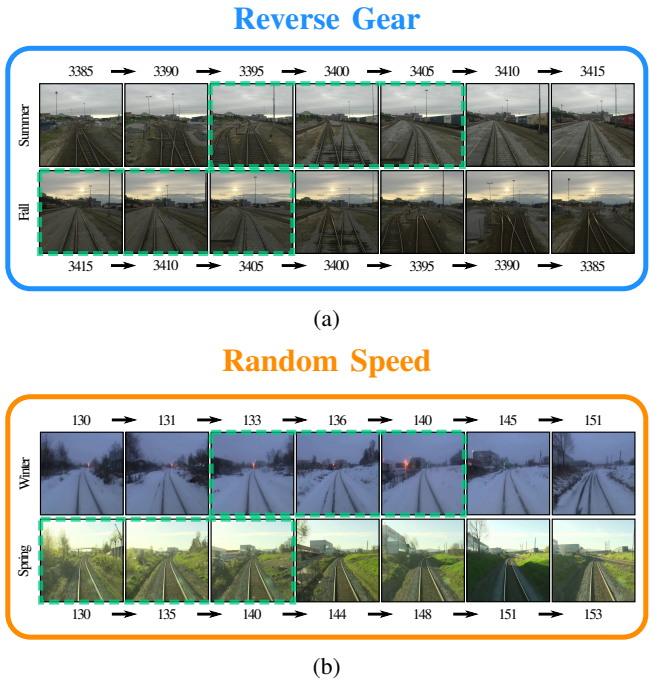

\begin{subfigure}[t]{0.99\linewidth}
    \centering
    \bf \large {\color{cRG}\experimentone{}}
\end{subfigure}
\medskip
\vspace{0.01cm}
\medskip
\begin{subfigure}[t]{0.99\linewidth}
    \includegraphics[width=\linewidth]{sections/figure/RG-ilovepdf-compressed.pdf}
   \subcaption{\label{fig:rg}}
\end{subfigure}

\begin{subfigure}[t]{0.99\linewidth}
    \centering
    \bf \large {\color{cRS}\experimenttwo{}}
\end{subfigure}
\medskip
\vspace{0.01cm}
\medskip
\begin{subfigure}[t]{0.99\linewidth}
    \includegraphics[width=\linewidth]{sections/figure/RS-ilovepdf-compressed.pdf}
   \subcaption{\label{fig:rs}}
\end{subfigure}

\caption{\label{fig:experiment_setup}\textbf{Experiment setup details. }(a) {\color{cRG}\textbf{\experimentone{}}}, in which the sequence is played in reverse order for one of the seasons (\textit{Fall} in the figure). (b) {\color{cRS}\textbf{\experimenttwo{}}}, in which the vehicle speed is modified for both seasons, reference (\textit{Winter}) and query (\textit{Spring}). In both cases, we mark with a{\color{greenbox}\textbf{ dashed green box}} the same-place three-frames sequences.}

\end{figure}
\begin{table}
\centering

\begin{tabular}{ccccc|c}
\textbf{Method} & \multicolumn{1}{p{1cm}}{\centering \textbf{Number\\of\\frames}}& {\color{cNT}\textbf{NT}} & {\color{cRG}\textbf{RG}} & {\color{cRS}\textbf{RS}} & \textbf{M/S} \\
\hline
    & $\#$& $\%$ & $\%$& $\%$& $\%$ \\
    
    Seqslam \cite{milford2012seqslam} & $3$ &   33\% &0.08\%&9\%&14.0$\pm$13.9\\
    Seqslam \cite{milford2012seqslam} & $10$&   70\%&0.03\%&8\%&26.01$\pm$31.27\\
    \textbf{ours (grouping)} & $3$ &   \textbf{92\%}& 74\% &36\%&67.3$\pm$23.3\\
    \textbf{ours (fusion)} & $3$ &   86\%&80\%&78\%&81.33$\pm$3.4\\
    \textbf{ours (recurrent)} & $3$ &  86\%&\textbf{82\%}&\textbf{84}\%&\textbf{84.0$\pm$1.6}\\

    \cline{3-5}
  \multicolumn{2}{c}{}&\multicolumn{3}{c}{biggest the best}
  
\end{tabular}
\caption{\label{tab:norland_rob} Experimental results for {\color{cRG}\textbf{\experimentone{}}} and {\color{cRS}\textbf{\experimenttwo{}}} in the \textbf{Partitioned Norland Dataset} \cite{olid2018single}. \textbf{\nth{1} column:} Method. \textbf{\nth{2} column:} Number of frames used for recognition. \textbf{\nth{3}-\nth{5} column:}  Summer vs Winter experiments: \textit{\nth{3} column:} {\color{cNT}\textbf{NT}} stands for ``Normal Test'' and corresponds to the one showed on Table \ref{tab:norland}. \textit{\nth{4} column:} {\color{cRG}\textbf{RG}} stands for ``Reverse Gear'' (the query frames are all in reversed order, \textit{i.e.} simulating the train has used a \textit{reverse gear}). {\color{cRS}\textbf{RS}} stands for ``Random Speed'' (the speed of the train is simulated to be random, which means some of the frames are lost). The speed variations are independent for the query and the reference databases, and this implies no more 1 to 1 correspondence.}
\end{table}
\noindent\textbf{Sequence Speed Changes:} Inspecting the previous results (Table \ref{tab:norland}), \modelone{} (\textbf{ours (grouping)}) trained only on single-view and then applied on multi-view by concatenation is the best performing. This is surprising at first sight, as the other two models were trained on multi-view data. We designed two extra experiments ({\color{cRG}\textbf{Reverse Gear}} and {\color{cRS}\textbf{Random Speed}}) to illustrate why this is happening. 

The {\color{cRG}\textbf{Reverse Gear}} experiment consist on changing the direction of the train motion on one of the sequences at test time (\textit{e.g.} when testing Winter vs Fall, the sequence of Fall is played in reverse order, see Fig \ref{fig:rg}). This experiment will help to discern how much the model exploits the multi-view information rather than just the sequence consistency. Table \ref{tab:norland_rob} shows that, as we expected, models trained with multi-view examples (\textbf{ours (fusion)} and \textbf{ours (recurrent)}) have learned to exploit multiple views: Its performance only degrades by $6\%$ and $4\%$ respectively. On the other side, \textbf{ours (grouping)} drops performance by $18\%$. 

In the {\color{cRS}\textbf{Random Speed}} 
experiment we modified the speed of the train motion on one of the sequences at test time.
Specifically, we modified the frame rate along the sequence simulating changes on the train velocity, see Fig. \ref{fig:rs} (in our experiments the velocity was randomly multiplied by $\times1$, $\times2$ or $\times3$ at every moment of the sequence). The ``speed'' is modified for the whole sequence, implying that the one-to-one correspondence in plain Nordland does not hold. Table \ref{tab:norland_rob} proves that {\color{cRS}\textbf{Random Speed}} is the most challenging setup for the \textbf{ours (grouping)} approach, dropping its precision to $36\%$. \textbf{ours (fusion)} and \textbf{ours (recurrent)} keep its performance at a very similar level than the standard Nordland setup ($78\%$ and $84\%$ respectively). 

We run these {\color{cRG}\textbf{RG}} and {\color{cRS}\textbf{RS}} experiments using the state-of-the-art multi-view baseline SeqSLAM \cite{milford2012seqslam}, observing that its performance drops in both. This should be expected, as SeqSLAM assumes a linear relation between the velocities of the query and the reference sequences (sequence consistency). 

The last column of Table \ref{tab:norland_rob} (\textbf{M/S}) summarizes the conclusions of both experiments, reporting the mean (biggest the best) and standard deviation (smallest the best) for all experiments ({\color{cNT}\textbf{NT}}, {\color{cRG}\textbf{RG}} and {\color{cRS}\textbf{RS}}). Observe that \textbf{ours (recurrent)} is the best performing, presenting both the highest average precision and smallest variations. This confirms our hypothesis: The sequence descriptors that incorporate learning to combine single-view features (\textbf{ours (fusion)} and (\textbf{ours (recurrent)}) are more resilient than those based on plain concatenation (\textbf{ours (grouping)}) or handcrafted relations (SeqSLAM).

\subsection{Alderley}
\begin{figure}[t!]

\includegraphics[width=\linewidth]{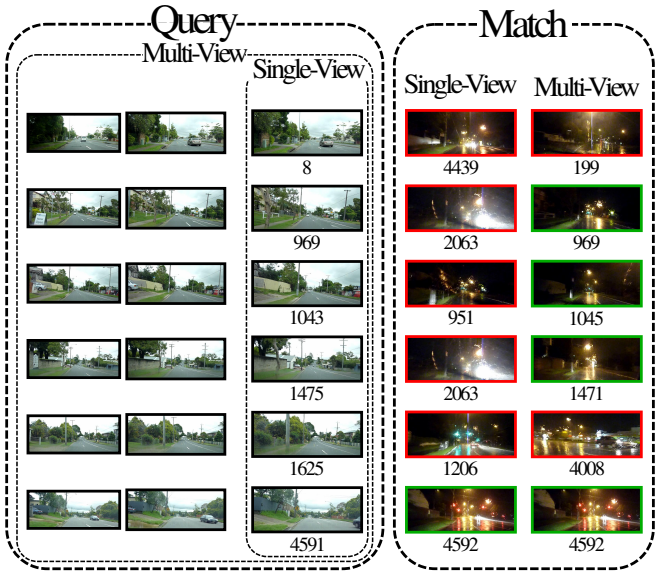}

\caption{ \label{fig:alder_qual}\textbf{Examples of Matched Places} for Single and the grouping Multi-View model in the Alderley Dataset \cite{milford2012seqslam}. The returned image is framed on {\color{green} green} if it is a correct match or on{\color{red}red} if it is incorrect.}

\end{figure}
\begin{table}
\centering

\begin{tabular}{ccccc}
\textbf{Method} &\multicolumn{1}{p{0.8cm}}{\centering\textbf{Trained\\on}} & \multicolumn{1}{p{0.9cm}}{\centering\textbf{Number\\of\\frames}}&\multicolumn{1}{p{0.9cm}}{\centering\textbf{Descriptor\\Size}}& D vs N  \\
\hline
    & & $\#$& & $\%$  \\
    Olid {\em et~al.} \cite{olid2018single}& Norland& $1$& 128 & 0.15\%\\
    Olid {\em et~al.} \cite{olid2018single}&Alderley& $1$& 128 & 6.84\%\\
    \textbf{ours (single-view)} & Norland& $1$& 128 & 1.65\%\\
    \textbf{ours (single-view)} &Alderley& $1$& 128 & 6.8\%\\
    \hline
    Seqslam \cite{milford2012seqslam} &-& $3$&6144 & 3.91\%\\
    Seqslam \cite{milford2012seqslam} &-& $10$& 20480& \textbf{9.90\%}\\
    \textbf{ours (grouping)}& Norland & $3$& 384 & 1.73\%\\
    \textbf{ours (grouping)} & Alderley& $3$& 384 & \textbf{7.82\%}\\
    \cline{5-5}
  \multicolumn{4}{c}{}&\multicolumn{1}{c}{biggest the best}
  
\end{tabular}
\caption{\label{tab:alderley} Results on the \textbf{Alderley Dataset} \cite{milford2012seqslam}, \textbf{\nth{1} column:} Method. \textbf{\nth{2} column:} Dataset in which the model it has been trained with. \textbf{\nth{3} column:} Number of frames used for recognition (\textit{e.g.} 1 would imply to be single-view).\textbf{ \nth{4} column:} Descriptor size in 32b floating point numbers.\textbf{\nth{5} column:} Day vs Night (D vs N), query with daylight image while reference database composed by nighttime images.}
\end{table}
We also evaluated our approach on the Alderley dataset \cite{milford2012seqslam}, that contains $15K$ images of a car trip in the day, and the same trip at night. It is a very challenging dataset due to the extreme illumination changes. We used the last $4600$ images as test samples. As the car velocity is similar in both cases, we only evaluate \textbf{ours (grouping)}.

Table \ref{tab:alderley} shows that our multi-view approach is the best performing model, outperforming the rest both in precision and descriptor compacity. We also compare the difference when training on the Partitioned Norland Dataset or the Alderley Dataset. Fine-tuning on Alderley clearly helps on the task, as the variant conditions (\textit{seasons} vs \textit{day/night}) impact differently in the visual appearance.

\textbf{Qualitative results}. Fig. \ref{fig:alder_qual} shows several test samples. Notice the increased challenge with respect to the Nordland dataset, with the presence of severe illumination changes plus inclusion of artificial illumination and dynamic objects.

\subsection{Execution time}

\begin{table}
\centering

\begin{tabular}{ccc|c}
\textbf{Method}& \multicolumn{1}{p{1.1cm}}{\centering\textbf{Descriptor\\Size}}& \multicolumn{1}{p{1.3cm}}{\centering \textbf{Descriptor\\Extraction}}& \multicolumn{1}{p{1.3cm}}{\centering \textbf{Search\\\textit{1 vs 10K}}}  \\
    \hline
   & & $ms$ & $ms$ \\
    
    \textbf{ours (fusion)} &128&17&\textbf{3.86}\\
    \textbf{ours (recurrent)}&128&22&\textbf{3.86}\\
    \textbf{ours (grouping)} &384&\textbf{15}&10.70\\
    \hline
                        -  &1860&-&49.21\\
                        -  &4096&-&111.44\\
                        -  &6144&-&166.13\\
                        -  &20480&-&688.24\\
                        -  &204800&-&9279.62\\
    \cline{3-4}
  \multicolumn{2}{c}{}&\multicolumn{2}{c}{smallest the best}
  
\end{tabular}
\caption{\label{tab:times} \textbf{Execution Time} of all our models. \textbf{\nth{1} column:} Method. \textbf{\nth{2} column:} Descriptor size. \textbf{\nth{3} column:} Time in milliseconds needed to extract 1 descriptor. \textbf{\nth{4} column:} Given a descriptor and a reference data base of $10K$ descriptors, time in milliseconds  needed to find the best match. }
\end{table}

We compared the execution time of all our models in the upper part of  Table \ref{tab:times}. The fourth column (\textbf{Descriptor Extraction}) shows the time needed to extract the descriptor of a query 3-frames sequence on a NVIDIA TITAN Xp. In this part of our pipeline, \modelone{} proved to be the fastest method as is uses the simplest network.

Last column (\textbf{Search}) shows the time needed to find the best match (Nearest Neighbor (NN)) given a query and a database of $10K$ examples. Notice that in this second part of our pipeline our methods \modeltwo{} and \modelthree{} run faster. This was expected, as their descriptor sizes are $n$ (\quseqe{} size) times smaller (3 times in our experiments). 
Our NN algorithm consists on an exhaustive search through the database. We iterate over all the visited places, compute the distance between their descriptor and the new query and select the minimum. For the distance function we compute the \textit{Squared Euclidean Distance} ($d^2(\mathbf{d}_q,\mathbf{d}_i)$). The computational complexity of this search is  $\mathcal{O}(N)$ where $N$ is the number of elements in the database and for the distance function $\mathcal{O}(k)$ where $k$ is the descriptor size. It makes a total of $\mathcal{O}(kN)$ which is practically $\mathcal{O}(N)$ when $k\ll N$ or worse than $\mathcal{O}(N^2)$ when $k\gg N$. 

Additionally, we computed the search time corresponding to the sizes of some of the other descriptors used in Tables \ref{tab:norland} and \ref{tab:alderley} (bottom part of the table). As expected, the time increases with the descriptor size. Notice that high dimensional descriptors rule out the use of more efficient data structures, such as \textit{KD-trees}, to speed up the search techniques, since it is not possible to reject candidates by using the difference of a single coordinate \cite{marimont1979nearest}.

\section{Conclusions}
\label{sec:conclusions}

In this work we have introduced three deep learning-based multi-view place recognition models, that outperform existing baselines both in accuracy and compacity. We analyzed different approaches to combine the information of the features from multiple views (grouping, fusion and recurrent), and we evaluated them on different experimental setups in two public datasets: Partitioned Norland and Alderley. Each one of the models we propose has its own strengths and weaknesses. On the one side, \modelone{} ensures the sequential consistency of the frame in a sequence, achieving the best performance in the standard Nordland/Alderley benchmarks, where the interframe motion is similar in different runs. On the other side, \modeltwo{} and \modelthree{} are able to learn more complex relations between frames and hence proved to be better in cases where the velocities differ or the frames ordering is different. We also evaluated the computational complexity of the approaches, demonstrating its potential for robotic applications.

{

\bibliographystyle{IEEEtran}
\bibliography{IEEEabrv,mybibfile}
}

\end{document}